\documentclass[10pt,journal,cspaper,compsoc]{IEEEtran}

\usepackage{amsmath,amssymb}
\usepackage{algorithm}
\usepackage{algorithmic}
\usepackage{multirow}

\usepackage[aboveskip=8pt]{caption}

\usepackage[dvips]{graphicx}
\DeclareGraphicsExtensions{.pdf}
\usepackage[american]{babel}

\usepackage{tabularx}

\usepackage[bookmarks=false,colorlinks=true,linkcolor=black,citecolor=black,filecolor=black,urlcolor=black]{hyperref}
\usepackage{url}
\usepackage{cvpr}
\usepackage{multicol}

\usepackage{soul}
\usepackage[normalem]{ulem}
\usepackage{color}

\newcommand{\tn}[1]{\footnotesize{#1}}
\newcolumntype{x}{>\small c}

\newcommand{\tabincell}[2]{\begin{tabular}{@{}#1@{}}#2\end{tabular}}

\begin{document}

\title{Object Detection Networks on \\ Convolutional Feature Maps}

\author{Shaoqing~Ren,
        Kaiming~He,
        Ross Girshick,
        Xiangyu~Zhang,
        and~Jian~Sun
\IEEEcompsocitemizethanks{
\IEEEcompsocthanksitem The majority of this work was done when the authors were with Microsoft Research.
\IEEEcompsocthanksitem S. Ren is with University of Science and Technology of China.
\IEEEcompsocthanksitem K. He and R.~Girshick are with Facebook AI Research.
\IEEEcompsocthanksitem X. Zhang is with Xi'an Jiaotong University.
\IEEEcompsocthanksitem J.~Sun is with Megvii.
}
}

\IEEEcompsoctitleabstractindextext{%
\begin{abstract}
Most object detectors contain two important components: a feature extractor and an object classifier.
The feature extractor has rapidly evolved with significant research efforts leading to better deep convolutional architectures. The object classifier, however, has not received much attention and many recent systems (like SPPnet and Fast/Faster R-CNN) use simple multi-layer perceptrons. This paper demonstrates that carefully designing deep networks for object classification is just as important. We experiment with region-wise classifier networks that use shared, region-independent convolutional features. We call them ``Networks on Convolutional feature maps'' (NoCs).
We discover that aside from deep feature maps, a \emph{deep} and \emph{convolutional} per-region classifier is of particular importance for object detection, whereas latest superior image classification models (such as ResNets and GoogLeNets) do not directly lead to good detection accuracy without using such a per-region classifier. We show by experiments that despite the effective ResNets and Faster R-CNN systems, the design of NoCs is an essential element for the 1st-place winning entries in ImageNet and MS COCO challenges 2015.
\end{abstract}
% Note that keywords are not normally used for peer review papers.
}

% make the title area
\maketitle

\IEEEpeerreviewmaketitle

\vspace{-0.7cm}
\section{Introduction}

Most object detectors contain two important components: a feature extractor and an object classifier.
The feature extractor in traditional object detection methods is a hand-engineered module, such as HOG \cite{Dalal2005}.
The classifier is often a linear SVM (possibly with a latent structure over the features) \cite{Felzenszwalb2010}, a non-linear boosted classifier \cite{Wang2013}, or an additive kernel SVM \cite{Uijlings2013}.

Large performance improvements have been realized by training deep ConvNets \cite{Krizhevsky2012} for object detection. R-CNN \cite{Girshick2014}, one particularly successful approach, starts with a pre-trained ImageNet \cite{Deng2009} classification network and then fine-tunes the ConvNet, end-to-end, for detection.
Although the distinction between the feature extractor and the classifier becomes blurry, a logical division can still be imposed.
For example, an R-CNN can be thought of as a convolutional feature extractor, ending at the last pooling layer, followed by a multi-layer perceptron (MLP) classifier. This methodology, however, appears rather different from traditional methods.

A research stream \cite{Savalle2014,Girshick2015,Wan2015,Zou2014} attempting to bridge the gap between traditional detectors and deep ConvNets creates a hybrid of the two: the feature extractor is ``upgraded'' to a pre-trained deep ConvNet, but the classifier is left as a traditional model, such as a DPM \cite{Savalle2014,Girshick2015,Wan2015} or a boosted classifier \cite{Zou2014}. These hybrid approaches outperform their HOG-based counterparts \cite{Felzenszwalb2010,Wang2013}, but still lag far behind R-CNN, even when the hybrid model is trained end-to-end \cite{Wan2015}.
Interestingly, the detection accuracy of these hybrid methods is close to that of R-CNN when using a linear SVM on the last convolutional features, without using the multiple fully-connected layers\footnote{The mAP on PASCAL VOC 2007 is 45-47\% \cite{Savalle2014,Girshick2015,Wan2015,Zou2014} for hybrid methods, and is 47\% for R-CNN that just uses SVM on the last convolutional layer. Numbers are based on AlexNet \cite{Krizhevsky2012}.}.

The SPPnet approach \cite{He2014} for object detection occupies a middle ground between the hybrid models and R-CNN.
SPPnet, like the hybrid models but \emph{unlike} R-CNN, uses convolutional layers to extract full-image features. These convolutional features are independent of region proposals and are shared by all regions, analogous to HOG features.
For classification, SPPnet uses a region-wise MLP, just like R-CNN but \emph{unlike} hybrid methods.
SPPnet is further developed in the latest detection systems including Fast R-CNN \cite{Girshick2015a} and Faster R-CNN \cite{Ren2015}, which outperform the hybrid methods.

\begin{figure*}[t]
  \centering
  \includegraphics[width=0.8\linewidth]{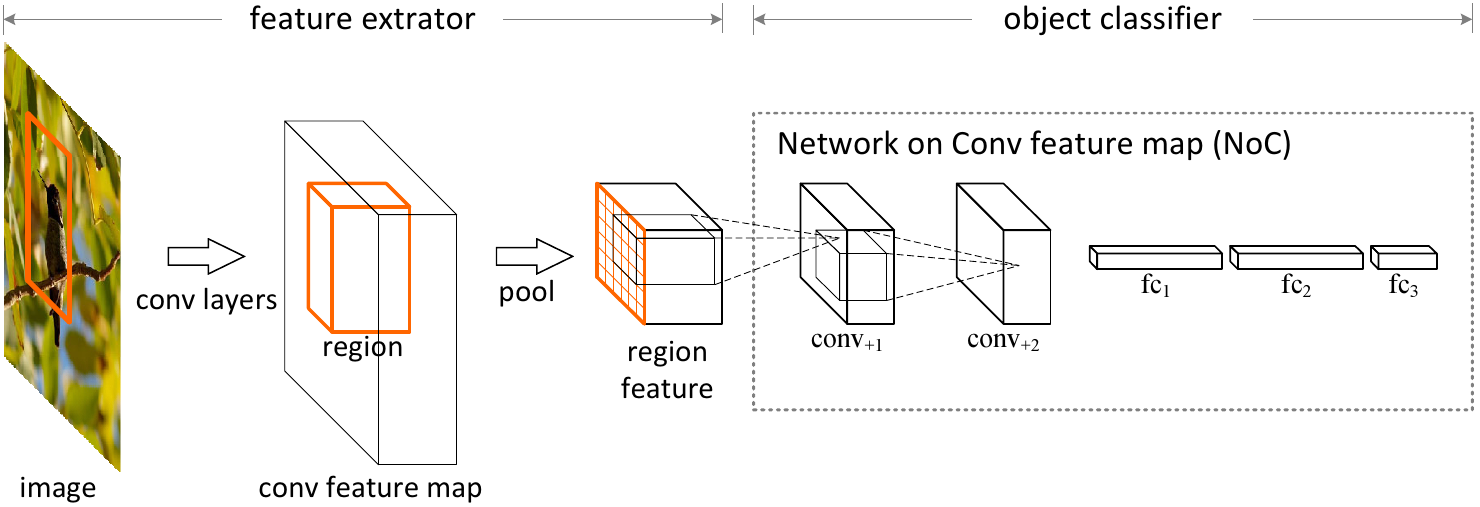}\\
  \caption{Overview of NoC. The convolutional feature maps are generated by the shared convolutional layers. A feature map region is extracted and RoI-pooled into a fixed-resolution feature. A new network, called a NoC, is then designed and trained on these features. In this illustration, the NoC architecture consists of two convolutional layers and three fully-connected layers.}\label{fig:overview}
  \vspace{-0.6cm}
\end{figure*}

From these systems \cite{He2014,Girshick2015a,Ren2015}, a prevalent strategy for object detection is now: use convolutional layers to extract \emph{region-independent} features, followed by \emph{region-wise} MLPs for classification. This strategy was, however, historically driven by pre-trained classification architectures similar to AlexNet \cite{Krizhevsky2012} and VGG nets \cite{Simonyan2015} that end with MLP classifiers.

In this paper, we provide an in-depth investigation into object detection systems from the perspective of classifiers aside from features. We focus on \emph{region-wise} classifier architectures that are on top of the shared, region-independent convolutional features. We call them ``Networks on Convolutional feature maps'', or \emph{NoCs} for short.
Our study brings in new insights for understanding the object detection systems.

Our key observation is that carefully designed region-wise classifiers improve detection accuracy over what is typically used (MLPs). We study three NoC families: MLPs of various depths, ConvNets of various depths, and ConvNets with maxout \cite{Goodfellow2013} for latent scale selection, where the latter two are unexplored families in previous works \cite{He2014,Girshick2015a,Ren2015}.
Ablation experiments suggest that: (i) a \emph{deep} region-wise classifier is important for object detection accuracy, in addition to deep shared features;
(ii) \emph{convolutional} layers for extracting \emph{region-wise} features are effective, and are complementary to the effects for extracting full-image shared features.

Based on these observations, we present an effective way of plugging ``fully convolutional'' image classifiers (such as ResNets~\cite{He2015a} and GoogLeNets~\cite{Szegedy2014}) into the Faster R-CNN \cite{Ren2015} system that was designed for the ``semi-convolutional'' VGG nets \cite{Simonyan2015}. We report that \emph{superior image classification backbones (\eg, ResNets and GoogLeNets) do not directly lead to better object detection accuracy}, and a \emph{deep}, \emph{convolutional} NoC is an essential element for outstanding detection performance, in addition to Faster R-CNN and extremely deep ResNets (more details in Table~\ref{tab:fasterrcnn_coco}).

In summary, through NoC we investigate the region-wise classifiers from different aspects, which are orthogonal to the investigation of features. We believe the observations in this paper will improve the understandings of ConvNets for object detection and also boost the accuracy of prevalent detectors such as Faster R-CNN \cite{Ren2015}.

\section{Related Work}

\noindent\textbf{Traditional Object Detection.}
Research on object detection in general focuses on both features and classifiers.
The pioneering work of Viola and Jones \cite{Viola2001} uses simple Haar-like features and boosted classifiers on sliding windows. The pedestrian detection method in \cite{Dalal2005} proposes HOG features used with linear SVMs. The DPM method \cite{Felzenszwalb2010} develops deformable graphical models and latent SVM as a sliding-window classifier. The Selective Search paper \cite{Uijlings2013} relies on spatial pyramid features \cite{Lazebnik2006} on dense SIFT vectors \cite{Lowe2004} and an additive kernel SVM. The Regionlet method \cite{Wang2013} learns boosted classifiers on HOG and other features.

\vspace{6pt}
\noindent\textbf{ConvNet-based Object Detection.}
Convolutional layers can be applied to images of arbitrary size yielding proportionally-sized feature maps. In the Overfeat method \cite{Sermanet2014}, the fully-connected layers are used on each sliding window of the convolutional feature maps for efficient classification, localization, and detection. In the SPP-based object detection method \cite{He2014}, features are pooled from proposal regions \cite{Uijlings2013} on convolutional feature maps, and fed into the original fully-connected layers for classifying.

Concurrent with this work, several papers \cite{Girshick2015a,Ren2015,Lenc2015,Gidaris2015} improve on the SPPnet method, inheriting the same logical division of shared convolutional features and region-wise MLP classifiers. In Fast R-CNN \cite{Girshick2015a}, the shared convolutional layers are fine-tuned end-to-end through Region-of-Interest pooling layers. In Faster R-CNN \cite{Ren2015}, the shared features are also used for proposing regions and reducing the heavy proposal burdens. The ``R-CNN minus R'' method \cite{Lenc2015} waives the requirement of region proposal by using pre-defined regions in the SPPnet system. In the Multi-Region method \cite{Gidaris2015}, the features are pooled from regions of multiple sizes to train an ensemble of models.

Despite the improvements, these systems \cite{He2014,Girshick2015a,Ren2015,Lenc2015,Gidaris2015} all use MLPs as region-wise classifiers. This logical division naturally applies to a series of networks, such as AlexNet \cite{Krizhevsky2012}, Zeiler and Fergus's (ZF) net \cite{Zeiler2014}, OverFeat \cite{Sermanet2014}, and VGG nets \cite{Simonyan2015}, which all have multiple fine-tunable fc layers. But this is not the case for fully convolutional classification networks, \eg, ResNet \cite{He2015a} and GoogleNet \cite{Szegedy2014}, that have \emph{no hidden fully-connected (fc) layers}. We show that it is nontrivial for Fast/Faster R-CNN to achieve good accuracy using this type of networks.

\section{Ablation Experiments}
\label{sec:ablation}

Firstly we present carefully designed ablation experiments on the PASCAL VOC dataset \cite{Everingham2010}.
We note that experiments in this section are mainly designed based on the SPPnet system. Particularly, in this section we consider the following settings: (i) the shared feature maps are frozen (which are fine-tunable with Fast R-CNN \cite{Girshick2015a}) so we can focus on the classifiers; (ii) the proposals are pre-computed from Selective Search \cite{Uijlings2013} (which can be replaced by a Region Proposal Network (RPN) \cite{Ren2015}), and (iii) the training step ends with a post-hoc SVM (in contrast to the end-to-end softmax classifier in Fast R-CNN \cite{Girshick2015a}). We remark that observations in this section are in general valid when these restricted conditions are relaxed or removed \cite{Girshick2015a,Ren2015}, as shown in the next section with Faster R-CNN \cite{Ren2015}.

\vspace{6pt}
\noindent\textbf{Experimental Settings}

%\vspace{6pt}
%\noindent\textbf{Dataset.}
We experiment on the PASCAL VOC 2007 set \cite{Everingham2010}. This dataset covers 20 object categories, and performance is measured by mAP on the \emph{test} set of 5k images.
We investigate two sets of training images: (i) the original \emph{trainval} set of 5k images in VOC 2007, and (ii) an augmented set of 16k images that consists of VOC 2007 trainval images and VOC 2012 trainval images, following \cite{Agrawal2014}.

%\vspace{6pt}
%\noindent\textbf{Pre-trained Models.}
As a common practice \cite{Girshick2014,He2014}, we adopt deep CNNs pre-trained on the 1000-class ImageNet dataset \cite{Deng2009} as feature extractors. In this section we investigate Zeiler and Fergus's (ZF) model \cite{Zeiler2014} and VGG models \cite{Simonyan2015}. The ZF model has five convolutional (conv) layers and three fully-connected (fc) layers. We use a ZF model released by \cite{He2014}\footnote{\url{https://github.com/ShaoqingRen/SPP_net/}}. The VGG-16/19 models have 13/16 conv layers and three fc layers, released by \cite{Simonyan2015}\footnote{\url{www.robots.ox.ac.uk/~vgg/research/very_deep/}}.

\vspace{6pt}
\noindent\textbf{Outline of Method}

%\vspace{6pt}
%\noindent\textbf{Region Extraction.}
We apply the conv layers of a pre-trained model to compute the convolutional feature map of the entire image. As in \cite{He2014}, we extract feature maps from multiple image scales. In this section these pre-trained conv layers are frozen and not further tuned as in \cite{He2014}, so we can focus on the effects of NoCs.

We extract $\sim$2,000 region proposals by Selective Search \cite{Uijlings2013}. We pool region-wise features from the shared conv feature maps using Region-of-Interest (RoI) pooling \cite{Girshick2015a,He2014}. RoI pooling produces a \emph{fixed-resolution} ($m\times m$) feature map for each region, in place of the last pooling layer in the pre-trained model ($6\times6$ for ZF net and $7\times7$ for VGG-16/19). The pooled feature map regions can be thought of as tiny multi-channel images (see Fig.~\ref{fig:overview}).

%\vspace{6pt}
%\noindent\textbf{Training and Inference.}
We consider these $m\times m$-sized feature maps as a new data source and design various NoC architectures to classify these data. The NoC structures have multiple layers, and the last layer is an ($n$+1)-way classifier for $n$ object categories plus background, implemented by an ($n$+1)-d fc layer followed by softmax. Each NoC is trained by backpropagation and stochastic gradient descent (SGD). After network training, we use the second-to-last fc layer in the NoC to extract features from regions, and train a linear SVM classifier for each category using these features, for a fair comparison with \cite{Girshick2014,He2014}. The implementation details follow those in \cite{He2014}.

For inference, the RoI-pooled features are fed into the NoC till the second-to-last fc layer. The SVM classifier is then used to score each region, followed by non-maximum suppression \cite{Girshick2014}.

Next we design and investigate various NoC architectures as classifiers on the RoI-pooled features.

\subsection{Using MLP as NoC}

A simple design of NoC is to use fc layers only, known as a multi-layer perceptron (MLP) \cite{Hornik1989}.
We investigate using 2 to 4 fc layers. The last fc layer is always ($n$+1)-d with softmax, and the other fc layers are 4,096-d (with ReLU \cite{Nair2010}). For example, we denote the NoC structure with 3 fc layers as ``f4096-f4096-f21'' where 21 is for the VOC categories (plus background).

Table~\ref{tab:noc_mlp} shows the results of using MLP as NoC. Here we \emph{randomly initialize} the weights by Gaussian distributions.
The accuracy of NoC with 2 to 4 fc layers increases with the depth.
Compared with the SVM classifier trained on the RoI features (``SVM on RoI'', equivalent to a 1-fc structure), the 4-fc NoC as a classifier on the same features has 7.8\% higher mAP. Note that in this comparison the NoC classifiers have \emph{no pre-training} (randomly initialized). The gain is solely because that MLPs are better classifiers than single-layer SVMs.
In the special case of 3 fc layers, the NoC becomes a structure similar to the region-wise classifiers popularly used in SPPnet \cite{He2014} and Fast/Faster R-CNN \cite{Girshick2015a,Ren2015}.

\subsection{Using ConvNet as NoC}

\begin{figure}[t]
  \begin{center}
  \includegraphics[width=1.0\linewidth]{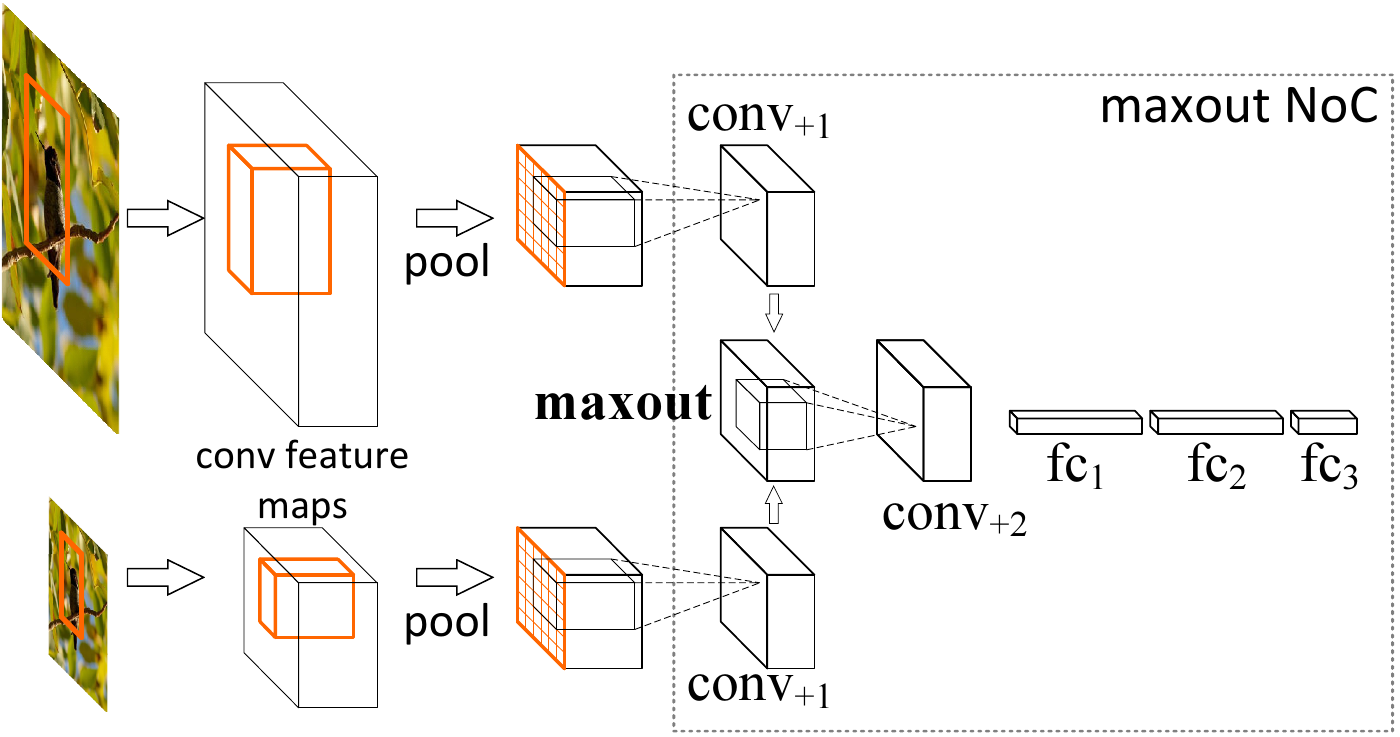}
  \end{center}
  \vspace{-0.1cm}
  \caption{A maxout NoC of ``c256-\textbf{mo}-c256-f4096-f4096-f21''. The features are RoI-pooled from two feature maps computed at two scales. In this figure, maxout is used after conv$_{+1}$.}
  \label{fig:maxout}
\end{figure}

\begin{table}[t]
\setlength{\tabcolsep}{8pt}
\renewcommand{\arraystretch}{1.1}
\small
\centering
\begin{tabular}{c|r|c}
 method & \multicolumn{1}{c|}{architecture} & \tn{VOC 07} \\
\hline
 SVM on RoI & \tn{f21} & 45.8 \\
\hline
2fc NoC & \tn{f4096-f21} & 49.0 \\
3fc NoC & \tn{f4096-f4096-f21} & 53.1 \\
4fc NoC & \tn{f4096-f4096-f4096-f21} &  \textbf{53.6} \\
\end{tabular}
\vspace{-0.2cm}
\caption{Detection mAP (\%) of \textbf{NoC as MLP} for PASCAL VOC 07 using a ZF net. The training set is PASCAL VOC 07 trainval. The NoCs are randomly initialized. No bbox regression is used.}
\label{tab:noc_mlp}
%\end{table}
\vspace{1em}
%\begin{table}[t]
\centering
\setlength{\tabcolsep}{3pt}
\renewcommand{\arraystretch}{1.1}
\small
\begin{tabular}{c|r|c|c}
 method & \multicolumn{1}{c|}{architecture} & \tn{VOC 07} & \tn{07+12} \\
\hline
3fc NoC & \tn{f4096-f4096-f21} & 53.1 & 56.5 \\
\hline
1conv3fc NoC & \tn{c256-f4096-f4096-f21} & \textbf{53.3} & 58.5\\
2conv3fc NoC & \tn{c256-c256-f4096-f4096-f21} & 51.4 & \textbf{58.9} \\
3conv3fc NoC & \tn{c256-c256-c256-f4096-f4096-f21} & 51.3 & 58.8 \\
\end{tabular}
\vspace{-0.2cm}
\caption{Detection mAP (\%) of \textbf{NoC as ConvNet} for PASCAL VOC 07 using a ZF net. The training sets are PASCAL VOC 07 trainval and 07+12 trainval respectively. The NoCs are randomly initialized. No bbox regression is used.}
\label{tab:noc_conv}
%\end{table}
\vspace{1em}
%\begin{table}[t]
\centering
\setlength{\tabcolsep}{6pt}
\renewcommand{\arraystretch}{1.1}
\small
\begin{tabular}{l|r|c}
 \multicolumn{1}{c|}{method} & \multicolumn{1}{c|}{architecture} & \tn{VOC 07+12} \\
\hline
 \multicolumn{1}{c|}{2conv3fc NoC} & \tn{c256-c256-f4096-f4096-f21} & 58.9 \\
\hline
mo input & \tn{\textbf{mo}-c256-c256-f4096-f4096-f21} & 60.1 \\
mo conv$_{+1}$ & \tn{c256-\textbf{mo}-c256-f4096-f4096-f21} & \textbf{60.7} \\
mo fc$_{1}$ & \tn{c256-c256-f4096-\textbf{mo}-f4096-f21} & 60.3\\
mo output & \tn{c256-c256-f4096-f4096-f21-\textbf{mo}} & 60.1\\
\end{tabular}
\vspace{-0.2cm}
\caption{Detection mAP (\%) of \textbf{maxout NoC} for PASCAL VOC 07 using a ZF net. The training set is 07+12 trainval. The NoCs are randomly initialized. No bbox regression is used.}
\label{tab:noc_maxout}
%\vspace{-.5em}
\vspace{-0.6cm}
\end{table}

In recent detection systems \cite{He2014,Girshick2015a,Ren2015,Lenc2015,Gidaris2015}, conv layers in the pre-trained models are thought of as \emph{region-independent} feature extractors, and thus are shared on the entire image \emph{without being aware of the regions that are of interest}. Although this is a computationally efficient solution, it misses the opportunities of using conv layers to learn \emph{region-aware} features that are fit to the regions of interest (instead of full images).
We investigate this issue from the NoC perspective, where the NoC classifiers may have their own conv layers.

We investigate using 1 to 3 additional conv layers (with ReLU) in a NoC.
We use 256 conv filters for the ZF net and 512 for the VGG net. The conv filters have a spatial size of 3$\times$3 and a padding of 1, so the $m\times m$ spatial resolution is unchanged. After the last additional conv layer, we apply three fc layers as in the above MLP case. For example, we denote a NoC with 2 conv layers as ``c256-c256-f4096-f4096-f21''.

In Table~\ref{tab:noc_conv} we compare the cases of no conv layer (3-layer MLP) and using 1 to 3 additional conv layers. Here we still randomly initialize all NoC layers.
When using VOC 07 trainval for training, the mAP is nearly unchanged when using 1 additional conv layer, but drops when using more conv layers. We observe that the degradation is a result of overfitting. The VOC 07 trainval set is too small to train deeper models.
However, NoCs with conv layers show improvements when trained on the VOC 07+12 trainval set (Table~\ref{tab:noc_conv}).
For this training set, the 3fc NoC baseline is lifted to 56.5\% mAP.
The advanced 2conv3fc NoC improves over this baseline to 58.9\%. This justifies the effects of the additional conv layers.
Table~\ref{tab:noc_conv} also shows that the mAP gets saturated when using 3 additional conv layers. %But it is reasonable to hope that more training data will further improve deeper convolutional NoCs, as we show in the next section on the MS COCO dataset.

Using a ConvNet as a NoC is not only effective for the ZF and VGG nets. In fact, as we show in the next section (Table~\ref{tab:fasterrcnn_coco}), this design is of central importance for Faster R-CNN using ResNets \cite{He2015a} and other fully convolutional pre-trained architectures.

\subsection{Maxout for Scale Selection}

Our convolutional feature maps are extracted from multiple discrete scales, known as a feature pyramid \cite{Felzenszwalb2010}.
In the above, a region feature is pooled from a single scale selected from the pyramid following \cite{He2014}.
Next, we incorporate a local competition operation (maxout) \cite{Goodfellow2013} into NoCs to improve scale selection from the feature pyramid.

To improve scale invariance, for each proposal region we select two adjacent scales in the feature pyramid. Two fixed-resolution ($m\times m$) features are RoI-pooled, and the NoC model has two data sources.
Maxout \cite{Goodfellow2013} (element-wise max) is a widely considered operation for merging two or multiple competing sources. %In DPM \cite{Felzenszwalb2010}, the prediction scores of multiple components compete with each other, and the component with the highest score is taken. This can be formulated as a maxout operation on the \emph{outputs} of multiple networks \cite{Girshick2015}. Alternatively, the maxout operation can be applied on the network \emph{inputs} from multiple scales to select the most responsive features.
%In the case of a deep structure, the maxout operation can also be applied on any intermediate layers.
We investigate NoCs with maxout used after different layers. For example, the NoC model of ``c256-mo-c256-f4096-f4096-f21'' is illustrated in Fig.~\ref{fig:maxout}. When the maxout operation is used, the two feature maps (for the two scales) are merged into a single feature of the same dimensionality using element-wise max.
There are two pathways before the maxout, and we let the corresponding layers in both pathways share their weights. Thus the total number of weights is unchanged when using maxout.

Table~\ref{tab:noc_maxout} shows the mAP of the four variants of maxout NoCs. Their mAP is higher than that of the non-maxout counterpart, by up to 1.8\% mAP. We note that the gains are observed for all variants of using maxout, while the differences among these variants are marginal.

\begin{table}[t]
\begin{center}
\setlength{\tabcolsep}{1pt}
\renewcommand{\arraystretch}{1.1}
\small
\begin{tabular}{c|c|c|c|c}
method & model & init. & \tn{VOC 07} & \tn{07+12} \\
\hline
SVM on RoI   & ZF &   -     & 45.8 & 47.7 \\
\hline
\multirow{2}{*}{3fc NoC} & \multirow{2}{*}{ZF} &random & 53.1 & 56.5 \\
                          & & pre-trained & \textbf{55.8} & \textbf{58.0} \\
\hline
\multirow{2}{*}{maxout 2conv3fc NoC}  & \multirow{2}{*}{ZF} & random & 54.7 & 60.7 \\
 & & pre-trained & \textbf{57.7} & \textbf{62.9} \\
\hline
\multirow{2}{*}{maxout 2conv3fc NoC}  & \multirow{2}{*}{\small{VGG-16}} & random & 59.4 & 65.0 \\
 & & pre-trained & \textbf{63.3} & \textbf{68.8} \\

\end{tabular}
\vspace{-0.2cm}
\end{center}
%\vspace{-.5em}
\caption{{Detection mAP (\%) of NoC for PASCAL VOC 07 using ZF/VGG-16 nets with different initialization. The training sets are PASCAL VOC 07 trainval and PASCAL VOC 07+12 trainval respectively. %The ``maxout 2conv3fc NoC'' is with maxout after conv$_{+1}$.
No bounding box regression is used.}}
\label{tab:noc_ft}
\end{table}

\subsection{Fine-tuning NoC}
\label{sec:finetune}

In the above, all NoC architectures are \emph{initialized randomly}. Whenever possible, we can still transfer weights from a pre-trained architecture and fine-tune the NoCs. The comparison of random initialization \vs fine-tuning provides new insights into the impacts of the well established fine-tuning strategy \cite{Girshick2014}.

For the fine-tuning version, we initialize the two 4096-d layers by the two corresponding fc layers in the pre-trained model. As such, the fine-tuned 3-fc NoC becomes equivalent to the SPPnet object detection system \cite{He2014}. For the cases of additional conv layers, each conv layer is initialized to the identity mapping, and thus the initial network state is equivalent to the pre-trained 3fc structure.
We compare the results of an SVM on RoI, randomly initialized NoC, and fine-tuned NoC initialized in the above way.
Table~\ref{tab:noc_ft} shows the cases of two NoCs.

Unsurprisingly, the fine-tuned models boost the results. However, it is less expected to see that the randomly initialized NoCs produce excellent results. Compared with the SVM counterpart using the same RoI-pooled features (47.7\%, Table~\ref{tab:noc_ft}), the randomly initialized NoC (60.7\%) showcases an improvement of 13.0\%, whereas the fine-tuned counterpart (62.9\%) has an extra 2.2\% gain. This indicates that the fine-tuning procedure, for the classifier, can obtain a majority of accuracy via training a \emph{deep} network on the detection data.

\subsection{Deep Features \vs Deep Classifiers}

\begin{table}[t]
\begin{center}
\setlength{\tabcolsep}{1pt}
\renewcommand{\arraystretch}{1.1}
\small
\begin{tabular}{c|c|c|c|c|c}
 & \footnotesize NoC & \footnotesize \tabincell{c}{ depth \\ (feature) } & \footnotesize \tabincell{c}{ depth \\ (classifier) } & \footnotesize \tabincell{c}{ depth \\ (total) } & \footnotesize mAP (\%) \\
\hline
VGG-16 & 3fc  & 13 & 3 & 16 & 64.6 \\
VGG-19 & 3fc  & 16 & 3 & 19 & 65.1 \\
VGG-16 & 2conv3fc  & 13 & 5 & 18 & \textbf{66.1}               \\
\hline
VGG-16 & maxout 2conv3fc  & 13 & 5 & 18 & \textbf{68.8} \\
\end{tabular}
\vspace{-0.2cm}
\end{center}
%\vspace{-.5em}
\caption{Detection results for PASCAL VOC 07 using VGG nets. The training set is PASCAL VOC 07+12 trainval. The NoC is the fine-tuned version (Sec.~\ref{sec:finetune}). %The ``maxout 2conv3fc NoC'' has maxout after conv$_{+1}$.
No bounding box regression is used.}
\label{tab:noc_vgg}
\vspace{-0.6cm}
\end{table}

\begin{figure}[t]
\begin{center}
  \includegraphics[height=0.47\columnwidth]{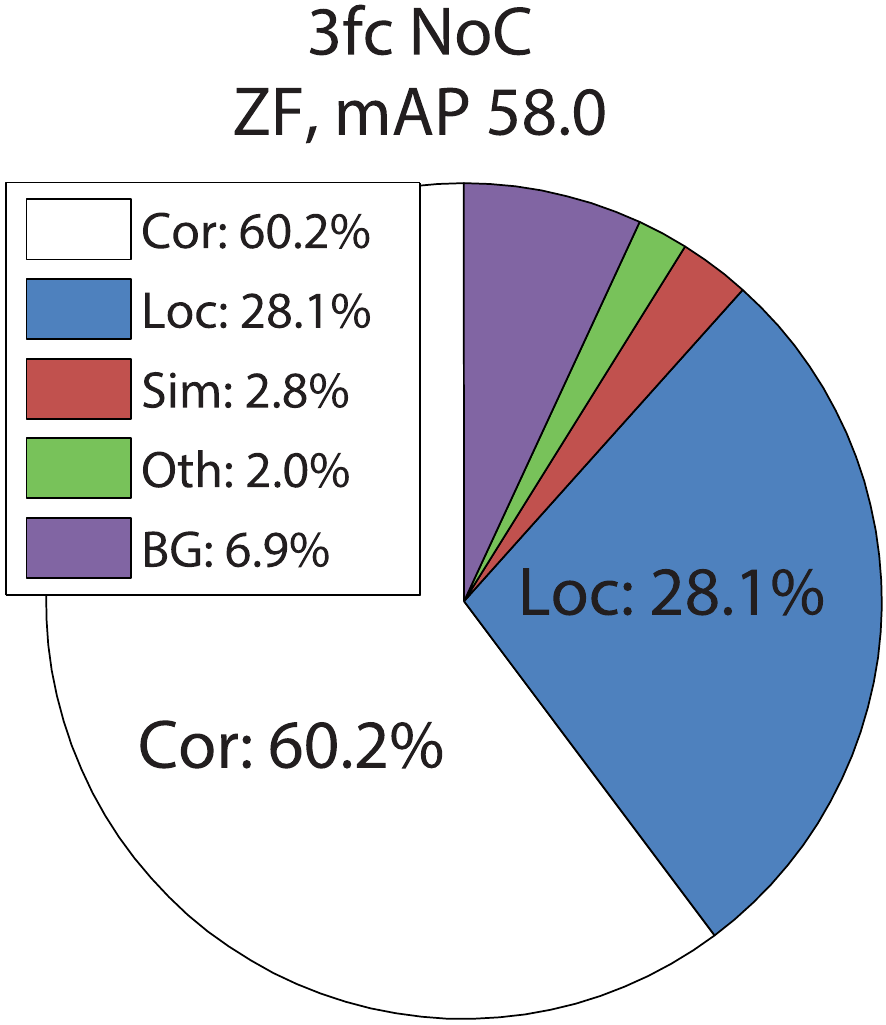}\quad\quad
  \includegraphics[height=0.47\columnwidth]{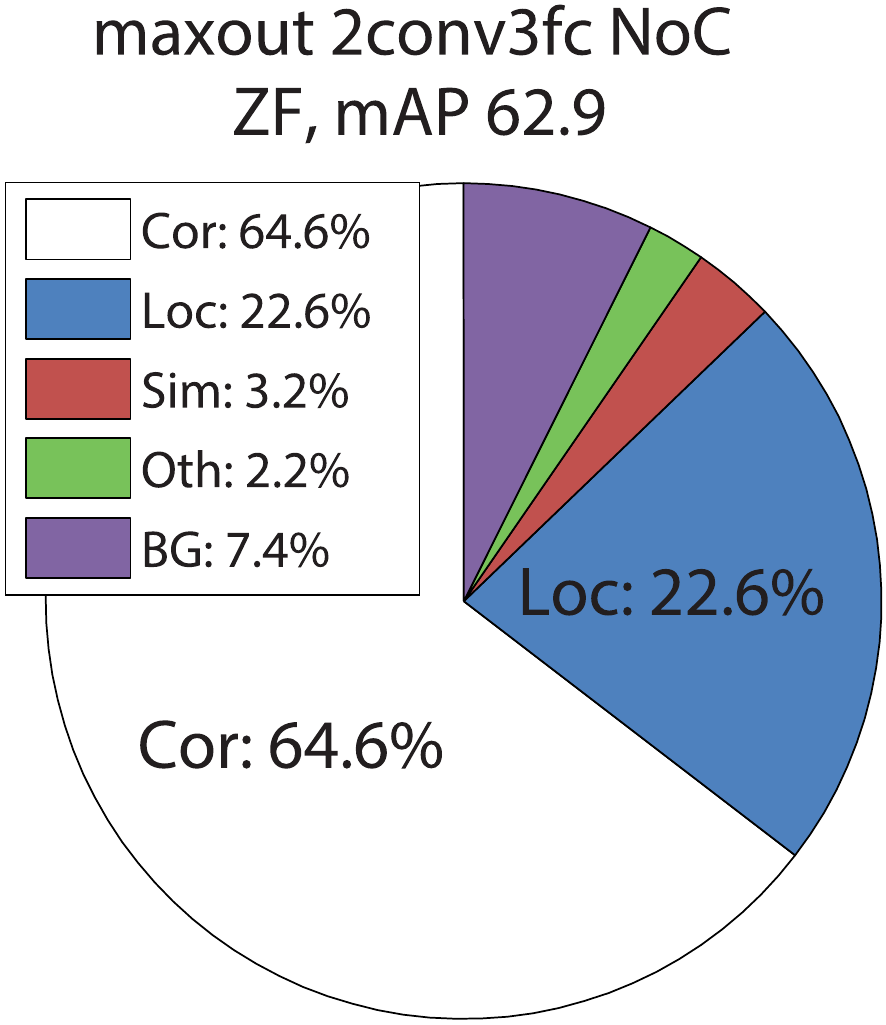}
  \\
  \vspace{1em}
  \includegraphics[height=0.47\columnwidth]{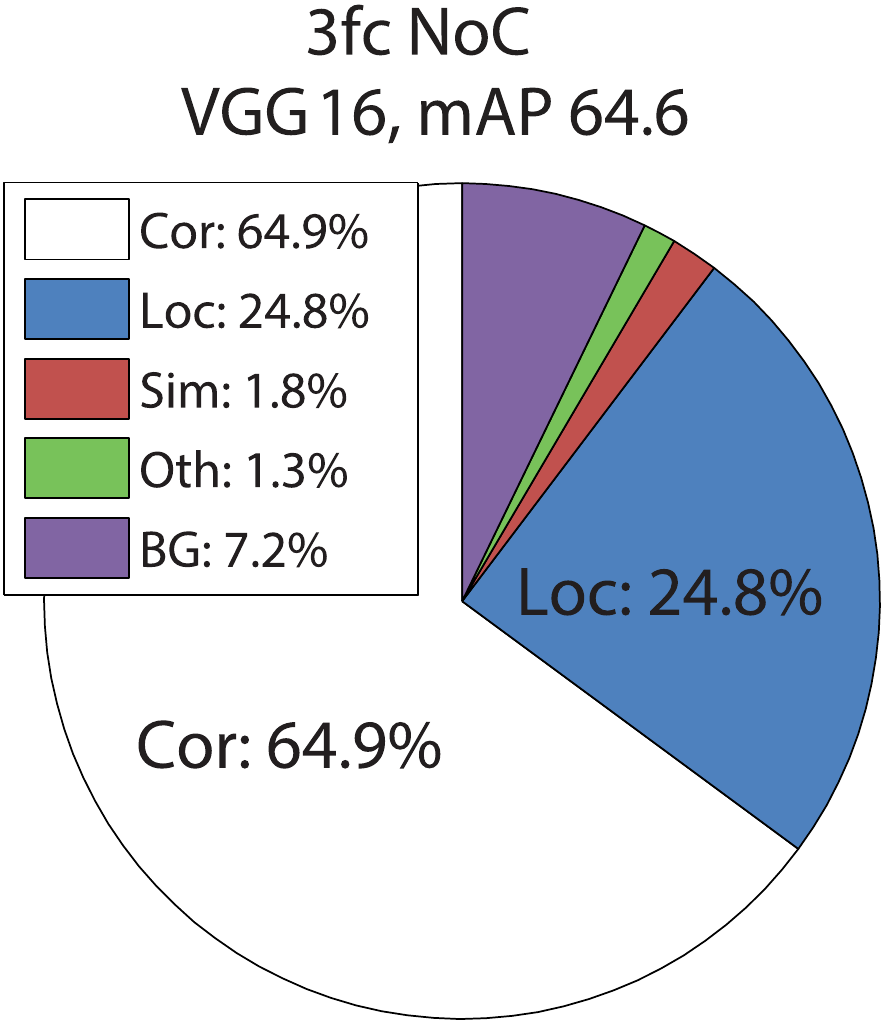}\quad\quad
  \includegraphics[height=0.47\columnwidth]{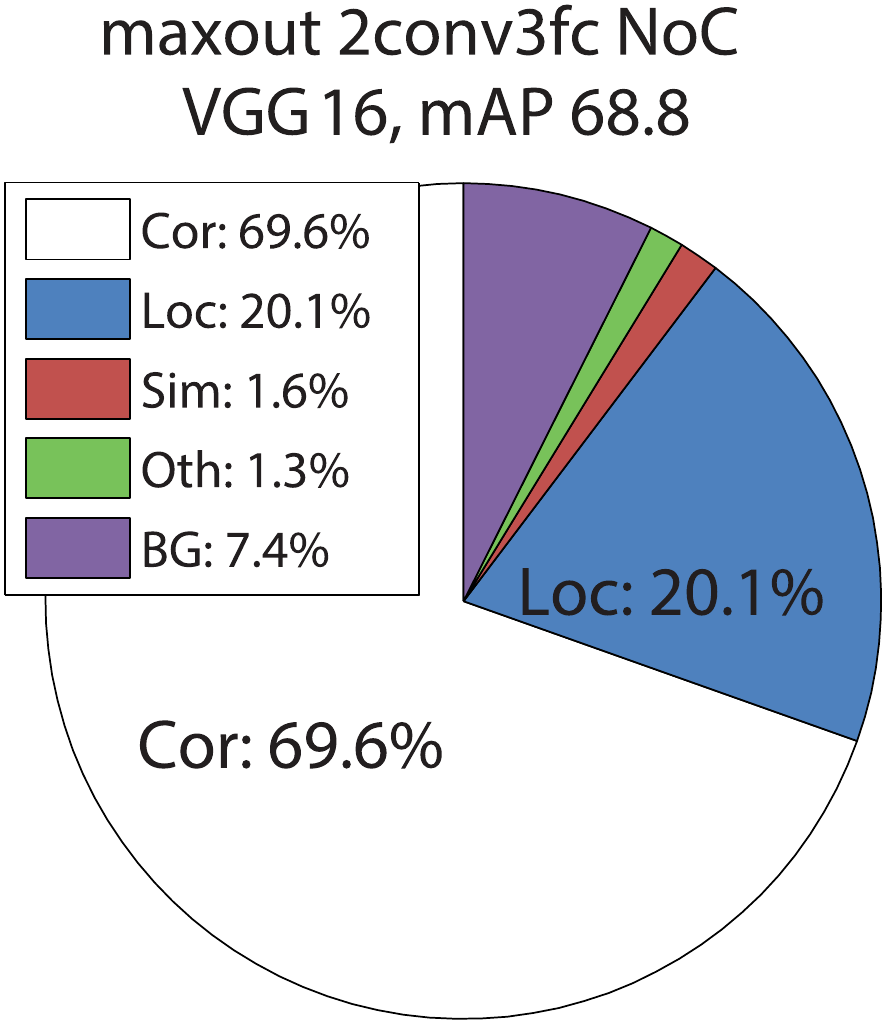}
\end{center}
\vspace{-0.1cm}
\caption{Distribution of top-ranked True Positives (TP) and False Positives (FP), generated by the published diagnosis code of \cite{Hoiem2012}.
The types of positive predictions are categorized \cite{Hoiem2012} as Cor (correct), Loc (false due to poor localization), Sim (confusion with a similar category), Oth (confusion with a dissimilar category), BG (fired on background).
The total number of samples in each disk is the same and equal to the total number of ground-truth labels \cite{Hoiem2012}.
More explanations are in the main text.}
\label{fig:error_analyses}
\vspace{-0.5cm}
\end{figure}

We further show by experiments that a deep classifier has \emph{complementary} effects to deep features.
Table~\ref{tab:noc_vgg} shows the NoC results using the VGG models \cite{Simonyan2015}. The mAP of the baseline 3fc NoC is 64.6\% with VGG-16. With the network replaced by the deeper VGG-19, the depth of shared features is increased by 3, and the mAP is increased by 0.5\% to 65.1\%. On the other hand, when the depth of region-aware classifier is increased (but still using the VGG-16 features), the mAP is increased by 1.5\% to 66.1\%. This means that for exploiting very deep networks, the depth of features and the depth of classifiers are both important.

%\vspace{6pt}
%\noindent\textbf{Error Analysis.}
\subsection{Error Analysis}
Our best NoC using VGG-16 has 68.8\% mAP (Table~\ref{tab:noc_vgg}). To separately investigate the gains that are caused by features (stronger pre-trained nets) and classifiers (stronger NoCs), in Fig.~\ref{fig:error_analyses} we analyze the errors of using two sets of pre-trained features (ZF \vs VGG-16) and two NoCs (3fc \vs maxout 2conv3fc). We use the diagnosis tool of \cite{Hoiem2012}.

The errors can be roughly decomposed into two parts: \emph{localization} error and \emph{recognition} error.
Localization error (``Loc'') is defined \cite{Hoiem2012} as the false positives that are correctly categorized but have no sufficient overlapping with ground truth. Recognition error involves confusions with a similar category (``Sim''), confusions with a dissimilar category (``Oth''), and confusions with background (``BG'').

Fig.~\ref{fig:error_analyses} shows that VGG-16 in general has lower \emph{recognition} error than the ZF net, when using the same classifiers (\eg, 1.6\%+1.3\%+7.4\% \vs 3.2\%+2.2\%+7.4\%). This suggests that the region-independent features perform more prominent for \emph{recognizing object categories}. On the other hand, when using a stronger NoC (maxout 2conv3fc), \emph{the localization error} is substantially reduced compared with the 3fc baseline (22.6\% \vs 28.1\% with ZF, and 20.1\% \vs 24.8\% with VGG-16). This suggests that the NoCs mainly account for localizing objects. This can be explained by the fact that localization-sensitive information is only extracted after RoI pooling and is used by NoCs.

\begin{table}[t]
\begin{center}
\setlength{\tabcolsep}{8pt}
\renewcommand{\arraystretch}{1.1}
\small
\begin{tabular}{l|c|c}
method & training data & mAP (\%) \\
\hline
R-CNN \cite{Girshick2014} & 07 & 62.2 \\
R-CNN \cite{Girshick2014} + bb & 07 & 66.0 \\
\hline
SPPnet \cite{He2014} & 07 & 60.4 \\
% SPPnet \cite{He2014} + bb & 07 & 63.1 \\
SPPnet \cite{He2014} & 07+12 & 64.6 \\
%SPPnet \cite{He2014} + bb & 07+12 &  \\
\hline
Fast R-CNN \cite{Girshick2015a} & 07+12 & 70.0 \\
\hline
Faster R-CNN \cite{Ren2015} & 07+12 & \textbf{73.2} \\
\hline
NoC [ours] & 07+12 & 68.8 \\
NoC [ours] + bb & 07+12 & \textbf{71.6} \\
\end{tabular}
\vspace{-0.2cm}
\end{center}
\caption{Detection results for the PASCAL VOC 2007 test set using the VGG-16 model \cite{Simonyan2015}. Here ``bb'' denotes post-hoc bounding box regression \cite{Girshick2014}.}
\label{tab:voc07}
%\end{table}
%\begin{table}[t]
\begin{center}
\setlength{\tabcolsep}{8pt}
\renewcommand{\arraystretch}{1.1}
\small
\begin{tabular}{l|c|c}
method & training data & mAP (\%) \\
\hline
R-CNN \cite{Girshick2014} & 12 & 59.2 \\
R-CNN \cite{Girshick2014} + bb & 12 & 62.4 \\
\hline
Fast R-CNN \cite{Girshick2015a} & 07++12 & 68.4 \\
\hline
Faster R-CNN \cite{Ren2015} & 07++12 & \textbf{70.4} \\
\hline
NoC [ours] & 07+12 & 67.6 \\
NoC [ours] + bb & 07+12 & \textbf{68.8} \\
\end{tabular}
\vspace{-0.2cm}
\end{center}
\caption{{Detection results for the PASCAL VOC 2012 test set using the VGG-16 model \cite{Simonyan2015}. Here ``bb'' denotes post-hoc bounding box regression \cite{Girshick2014}.}}
\label{tab:voc12}
\vspace{-0.5cm}
\end{table}

\subsection{Comparisons of Results}

In Table~\ref{tab:voc07} and Table~\ref{tab:voc12}, we provide system comparisons with recent state-of-the-art results, including R-CNN \cite{Girshick2014}, SPPnet \cite{He2014}, and the latest Fast/Faster R-CNN \cite{Girshick2015a,Ren2015} that are contemporary to this work. We note that all methods in Table~\ref{tab:voc07} and Table~\ref{tab:voc12} are based on Selective Search (SS) proposals \cite{Uijlings2013} ($\sim$2,000 regions per image), except for Faster R-CNN \cite{Ren2015} that uses learned proposals.

Our method achieves 71.6\% mAP on the PASCAL VOC 2007 test set. This accuracy is higher than Fast R-CNN \cite{Girshick2015a} that also uses SS proposals, and lower than Faster R-CNN \cite{Ren2015} that uses learned proposals.

Nevertheless, Fast/Faster R-CNN \cite{Girshick2015a,Ren2015} essentially applies a 3-fc NoC structure as the region-wise classifier, and thus the effect of NoCs is orthogonal to theirs. This effect is particularly prominent using the ResNets \cite{He2015a} as we show in the next section.

\subsection{Summary of Observations}

The following key observations can be concluded from the above subsections:

\vspace{.5em}

\textbf{(i)} A \textbf{deeper} region-wise classifier is useful and is in general orthogonal to deeper feature maps.

\textbf{(ii)} A \textbf{convolutional} region-wise classifier is more effective than an MLP-based region-wise classifier.

\vspace{.5em}

These observations are strongly supported by the experimental results on the more challenging MS COCO dataset (Table~\ref{tab:fasterrcnn_coco}), as we introduced in the next section.

\begin{table*}[t]
\vspace{6pt}
\begin{center}
\setlength{\tabcolsep}{8pt}
\renewcommand{\arraystretch}{1.2}
\small
\begin{tabular}{l|l|c|r|c|cc}
%        & & {trained in} \\
net & feature & stride & NoC & AP & AP@0.5 & AP@0.75 \\
\hline
\hline
VGG-16 & conv5$_3$ & 16 & fc$_{4096}$, fc$_{4096}$, fc$_{81}$ & 21.2 & 41.5 & 19.7\\
\hline\hline
GoogleNet & inc5b & 32 & fc$_{81}$ & 15.2 & 34.7 & 11.6 \\
GoogleNet & inc5b & 32 & fc$_{4096}$, fc$_{4096}$, fc$_{81}$ & 19.8 & 40.8 & 17.5 \\
\hline
GoogleNet & inc5b, \emph{\`{a} trous} & 16 & fc$_{81}$ & 18.6 & 39.4 & 15.8 \\
GoogleNet & inc5b, \emph{\`{a} trous} & 16 & fc$_{4096}$, fc$_{4096}$, fc$_{81}$ & 23.6 & 43.4 & 23.0 \\
GoogleNet & inc4d & 16 & inc4e,5a,5b, fc$_{81}$ & 24.8 & 44.4 & 25.2 \\
\hline\hline
ResNet-101 & res5c & 32 & fc$_{81}$ & 16.9 & 39.6 & 12.1 \\
ResNet-101 & res5c & 32 & fc$_{4096}$, fc$_{4096}$, fc$_{81}$ & 21.2  & 43.1  & 18.9 \\
\hline
ResNet-101 & res5c, \emph{\`{a} trous} & 16 & fc$_{81}$ & 21.3 & 44.4 & 18.3 \\
ResNet-101 & res5c, \emph{\`{a} trous} & 16 & fc$_{4096}$, fc$_{4096}$, fc$_{81}$ & 26.3  & 48.1  & 25.9 \\
ResNet-101 & res4b$_{22}$ & 16 & res5a,5b,5c, fc$_{81}$ & \textbf{27.2} & \textbf{48.4} & \textbf{27.6} \\
\end{tabular}
\vspace{-0.2cm}
\end{center}
\caption{Detection results of Faster R-CNN on the MS COCO val set. ``inc'' indicates an inception block, and ``res'' indicates a residual block.}
\label{tab:fasterrcnn_coco}
\vspace{-0.7cm}
\end{table*}

\section{NoC for Faster R-CNN with ResNet}
\label{sec:fasterrcnn}

The Fast/Faster R-CNN systems \cite{Girshick2015a,Ren2015} have shown competitive accuracy and speed using VGG nets.
For networks similar to ZF and VGG-16, Fast/Faster R-CNN are naturally applicable and their region-wise classifiers are 3fc NoCs. However, for ``fully convolutional'' models such as GoogleNets \cite{Szegedy2014} and ResNets \cite{He2015a}, there are no hidden fc layers for building region-wise classifiers.
\emph{We demonstrate that the NoC design is an essential factor for Faster R-CNN \cite{Ren2015} to achieve superior results using ResNets}.

\vspace{6pt}
\noindent\textbf{Experimental Settings}

In this section we experiment on the more challenging MS COCO dataset \cite{Lin2014} with 80 categories. We train the models on the 80k train set, and evaluate on the 40k val set. We evaluate both COCO-style AP (@ IoU $\in$ [0.5, 0.95]) as well as AP@0.5 and AP@0.75. We adopt the same hyper-parameters as in \cite{He2015a} for training Faster R-CNN on MS COCO.

We compare network architectures of VGG-16 \cite{Simonyan2015}, GoogleNet \cite{Szegedy2014}, and ResNet-101 \cite{He2015a}. The VGG-16 has center crop top-1 error of 28.5\% on the ImageNet classification val set.
Regarding GoogleNet, we train the BN-Inception model \cite{Ioffe2015} on ImageNet classification. Our reproduced GoogleNet has center crop top-1 error of 26.4\%, close to that reported in \cite{Ioffe2015} (25.2\%). The 101-layer ResNet is released by the authors of \cite{He2015a}, with center crop top-1 error of 23.6\%. Both GoogleNet and ResNet have no hidden fc layer, and instead end with global average pooling and a 1000-d classifier.

Unlike the above section that is based on the SPPnet framework, in this section we use the more advanced Faster R-CNN \cite{Ren2015} detector. The main differences are: (i) the entire networks including the features are fine-tuned end-to-end \cite{Girshick2015a}; (ii) the proposals are learned by a RPN \cite{Ren2015} with features shared; (iii) instead of post-hoc SVM, a softmax classifier and a jointly learned bounding box regressor \cite{Girshick2015a} are learned end-to-end. Nevertheless, these differences do not affect the design of the NoCs.

\vspace{6pt}
\noindent\textbf{Experimental Results}

Table~\ref{tab:fasterrcnn_coco} shows the results on MS COCO val. We discuss by diving the results into 3 cases as following.

\vspace{6pt}
\noindent\emph{\textbf{Na\"{\i}ve Faster R-CNN}}. By this we mean that the RoI pooling layer is na\"{\i}vely adopted after the last convolutional layer (conv5$_3$ for VGG-16, inc5b for GoogleNet, and res5c for ResNet). In all cases, we set the output resolution of RoI pooling as 7$\times$7. This is followed by a 81-d classifier (equivalent to a 1fc NoC).

Table~\ref{tab:fasterrcnn_coco} shows that VGG-16 has better AP (21.2\%) than both GoogleNet (15.2\%) and ResNet (16.9\%), even though VGG-16 has worse image-level classification accuracy on ImageNet.
One reason is that VGG-16 has a stride of 16 pixels on conv5$_3$, but GoogleNet and ResNet have a stride of 32 pixels on inc5b and res5c respectively. We hypothesize that a finer-resolution feature map (\ie, a smaller stride) contributes positively to object detection accuracy. To verify this, we reduce the stride of GoogleNet/ResNet from 32 to 16 by modifying the last stride=2 operation as stride=1. Then we adopt the ``hole algorithm'' \cite{Long2015,Chen2015} (``\emph{Algorithme \`{a} trous}'' \cite{Mallat1999}) on all following layers to compensate this modification. With a stride of 16 pixels, na\"{\i}ve Faster R-CNN still performs unsatisfactorily, with an AP of 18.6\% for GoogleNet and 21.3\% for ResNet.

We argue that this is because in the case of na\"{\i}ve Faster R-CNN, VGG-16 has a 3fc NoC but GoogleNet and ResNet has a 1fc NoC (Table~\ref{tab:fasterrcnn_coco}). As we observed in the above section, a \emph{deeper} region-wise NoC is important, even though GoogleNet and ResNet have deeper feature maps.

\vspace{6pt}
\noindent\emph{\textbf{Using MLP as NoC}}. Using the same settings of feature maps, we build a deeper MLP NoC by using 3 fc layers (f4096-f4096-fc81). As GoogleNet and ResNet have no pre-trained fc layers available, these layers are randomly initialized which we expect to perform reasonably (Sec.~\ref{sec:finetune}). This 3fc NoC significantly improves AP by about 4 to 5\% for ResNet (21.3\% to 26.3\% with a stride of 16, and 16.9\% to 21.2\% with a stride of 32). These comparisons justify the importance of a deeper NoC.

\vspace{6pt}
\noindent\emph{\textbf{Using ConvNet as NoC}}. To build a convolutional NoC, we move the RoI pooling layer from the last feature map to an intermediate feature map that has a stride of 16 pixels (inc4d for GoogleNet and res4b$_{22}$ for ResNet). The following convolutional layers (inc4e,5a,5b for GoogleNet and res5a,5b,5c for ResNet) construct the convolutional NoC. The \emph{\`{a} trous} trick is not necessary in this case.

With the deeper convolutional NoC, the AP is further improved, \eg, from 26.3\% to 27.2\% for ResNet. In particular, this NoC greatly improves \emph{localization} accuracy --- ResNet's AP@0.75 is increased by 1.7 points (from 25.9\% to 27.6\%) whereas AP@0.5 is nearly unchanged (from 48.1\% to 48.4\%).
This observation is consistent with that on PASCAL VOC (Fig.~\ref{fig:error_analyses}), where a deep convolutional NoC improves localization.

Table~\ref{tab:fasterrcnn_voccoco} shows the comparisons on PASCAL VOC for Faster R-CNN + ResNet-101. Both MLP and ConvNet as NoC (76.4\%) perform considerably better than the 1fc NoC baseline (71.9\%), though the benefit of using ConvNet as NoC is diminishing in this case.

\begin{table}[t]
\vspace{6pt}
\begin{center}
\setlength{\tabcolsep}{4pt}
\renewcommand{\arraystretch}{1.1}
\small
\begin{tabular}{l|r|c|c}
%        & & {trained in} \\
method & NoC & \tabincell{c}{AP on \\ COCO} & \tabincell{c}{mAP on \\ VOC07} \\
\hline
res5c, \emph{\`{a} trous} & fc$_{n+1}$ & 21.3 & 71.9 \\
res5c, \emph{\`{a} trous} & fc$_{4096}$, fc$_{4096}$, fc$_{n+1}$ & 26.3 & 76.4 \\
res4b$_{22}$ & res5a,5b,5c, fc$_{n+1}$ & \textbf{27.2} & \textbf{76.4} \\
\end{tabular}
\vspace{-0.2cm}
\end{center}
\caption{Detection results of Faster R-CNN + ResNet-101 on MS COCO val (trained on MS COCO train) and PASCAL VOC 2007 test (trained on 07+12), based on different NoC structures.}
\label{tab:fasterrcnn_voccoco}
\vspace{-0.6cm}
\end{table}

\vspace{6pt}
\noindent\textbf{Discussions}

The above system (27.2\% AP and 48.4\% AP@0.5) is the foundation of the detection system in the ResNet paper \cite{He2015a}. Combining with orthogonal improvements, the results in \cite{He2015a} secured the 1st place in MS COCO and ImageNet 2015 challenges.

The ablation results in Table~\ref{tab:fasterrcnn_coco} indicate that despite the effective Faster R-CNN and ResNet, it is not direct to achieve excellent object detection accuracy. In particular, a \emph{na\"{\i}ve} version of Faster R-CNN using ResNet has low accuracy (21.3\% AP), because its region-wise classifier is shallow and not convolutional. On the contrary, a \textbf{deep} and \textbf{convolutional} NoC is an essential factor for Faster R-CNN + ResNet to perform accurate object detection.

\section{Conclusion}

In this work, we delve into the detection systems and provide insights about the \emph{region-wise} classifiers. We discover that deep convolutional classifiers are just as important as deep convolutional feature extractors. Based on the observations from the NoC perspective, we present a way of using Faster R-CNN with ResNets, which achieves nontrivial results on challenging datasets including MS COCO.

%------------------------------------------------------------------------

\bibliographystyle{IEEEtran}
\bibliography{pami_noc_compact}

\end{document}